\documentclass[10pt,twocolumn,letterpaper]{article}

\usepackage{iccv}
\usepackage{times}
\usepackage{epsfig}
\usepackage{graphicx}
\usepackage{amsmath}
\usepackage{amssymb}
\usepackage{bm}
\usepackage{comment}
\usepackage{booktabs}
\usepackage{arydshln}
\usepackage{multirow}
\usepackage{subfigure}
\usepackage{bbm}
\usepackage{enumitem}
\usepackage[accsupp]{axessibility}

\usepackage[breaklinks=true,bookmarks=false]{hyperref}

\iccvfinalcopy 

\def\iccvPaperID{6083} 

\ificcvfinal\fi

\begin{document}

\title{GEDepth: Ground Embedding for Monocular Depth Estimation}

\author{Xiaodong Yang$^*$ \qquad Zhuang Ma$^*$ \qquad Zhiyu Ji$^*$ \qquad Zhe Ren\\
QCraft
}

\maketitle
\ificcvfinal\thispagestyle{empty}\fi

\def\thefootnote{*}\footnotetext{Authors contributed equally}

\begin{abstract}
Monocular depth estimation is an ill-posed problem as the same 2D image can be projected from infinite 3D scenes. Although the leading algorithms in this field have reported significant improvement, they are essentially geared to the particular compound of pictorial observations and camera parameters (i.e., intrinsics and extrinsics), strongly limiting their generalizability in real-world scenarios. To cope with this challenge, this paper proposes a novel ground embedding module to decouple camera parameters from pictorial cues, thus promoting the generalization capability. Given camera parameters, the proposed module generates the ground depth, which is stacked with the input image and referenced in the final depth prediction. A ground attention is designed in the module to optimally combine ground depth with residual depth. Our ground embedding is highly flexible and lightweight, leading to a plug-in module that is amenable to be integrated into various depth estimation networks. Experiments reveal that our approach achieves the state-of-the-art results on popular benchmarks, and more importantly, renders significant generalization improvement on a wide range of cross-domain tests. 
\end{abstract}


\section{Introduction}
\label{sec:intro}
Accurate depth acquisition is crucial for many robotics applications~\cite{luo2021simtrack,luo2021pillarmotion,tateno2017cnn,wang2019pseudo} as depth provides pivotal information for onboard tasks ranging from perception~\cite{li2023pillarnext}, prediction~\cite{wang2023prophnet} to planning~\cite{li2023tip}. Although range sensors (e.g., LiDAR) are widely used to produce precise depth measurements, there has been fast growing attention to camera based depth estimation from both academia and industry due to its portability and cost-effectiveness~\cite{bhat2021adabins, eigen2014depth, fu2018deep, yuan2022new}. A typical monocular depth estimation network adopts an encoder-decoder architecture, which can be trained in a supervised~\cite{eigen2014depth, fu2018deep} or self-supervised manner~\cite{godard2019digging, mahjourian2018unsupervised, zhou2017unsupervised}. Most of the existing works in this field focus on designing more advanced network architectures~\cite{bhat2021adabins,li2022depthformer} or engineering more effective loss functions~\cite{fu2018deep, long2021adaptive}.  
Another line of research estimates depth by exploiting stereo imagery~\cite{badki2020bi3d,kusupati2020normal}, which however requires multiple calibrated cameras and is relatively more expensive and complicated. Compared with its stereo counterpart, the monocular depth estimation hinging on single cameras is more amenable for real-world deployment such as autonomous driving~\cite{bevdepth,philion2020eccv}. 

\begin{figure}[t]
\centering
\includegraphics[width=0.89\linewidth]{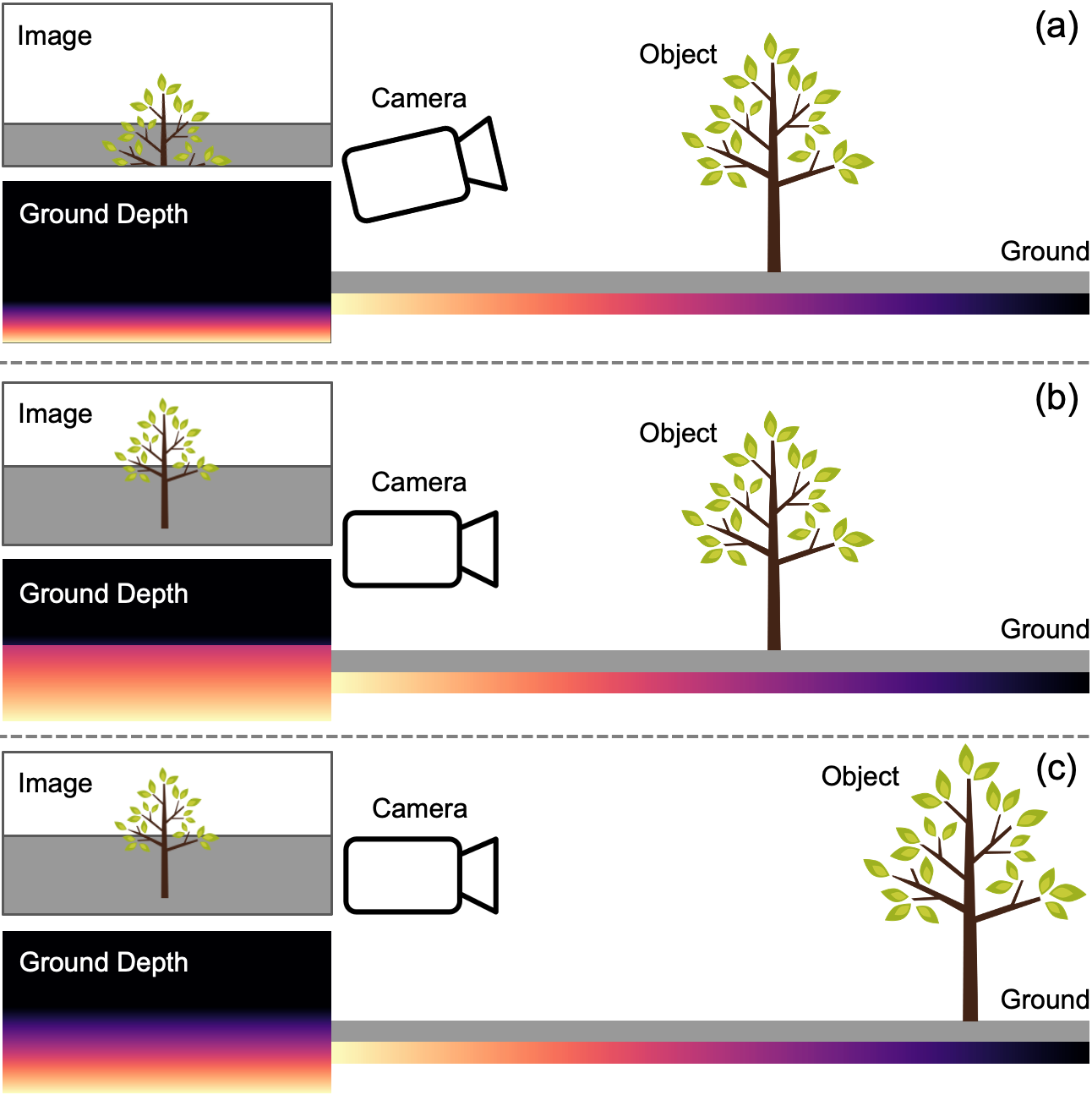}
\caption{Illustration of the ground depth produced by our ground embedding. Due to different camera parameters, (a) and (b) show one object with the same scale and depth is mapped to different images, while (b) and (c) show objects with different scales and depths generate the same image. In each case, the ground depth encodes the corresponding camera parameters and can be used to resolve the ill-posed problem in monocular depth estimation.   
}
\label{fig:teaser_figure}
\end{figure}

However, monocular depth estimation is inherently ill-posed or ambiguous. 
According to the classic pinhole camera model, an image captured by a camera is determined by the camera parameters (i.e., intrinsics and extrinsics), object scale, and object depth with respect to the camera optical center. Therefore, objects with different scales and depths captured by a camera could generate the same image; on the contrary, one object with the same depth captured by cameras with different intrinsics and or extrinsics could generate different images. As a result, one image may correspond to multiple plausible depths, and one depth can be mapped to various images, as illustrated in Figure~\ref{fig:teaser_figure}. Recently, most supervised methods resort to CNNs or Transformers to directly learn the absolute depth based on a single image from numerous labeled data~\cite{bhat2021adabins, fu2018deep, lee2019big, li2022depthformer, yuan2022new}. Although these methods have achieved remarkable progress, they are essentially tailored toward the specific compound of pictorial cues (e.g., object scale, light and shade, and texture gradient) and camera parameters on the particular training data, which strongly limits their generalization capability. In addition, very few works have provided an in-depth investigation into what these networks have learned to conduct monocular depth estimation, further undermining the guarantee of correct behaviors of these networks in cross-domain or unexpected scenarios. By inspecting synthetic images, \cite{dijk2019neural} shows that the ground contact point of an object is primarily used to estimate its depth. Another work~\cite{hu2019visualization} finds that the region around vanishing point of a scene contributes to the vital cues for depth estimation. 

\begin{figure*}[t]
\centering
\includegraphics[width=0.94\linewidth]{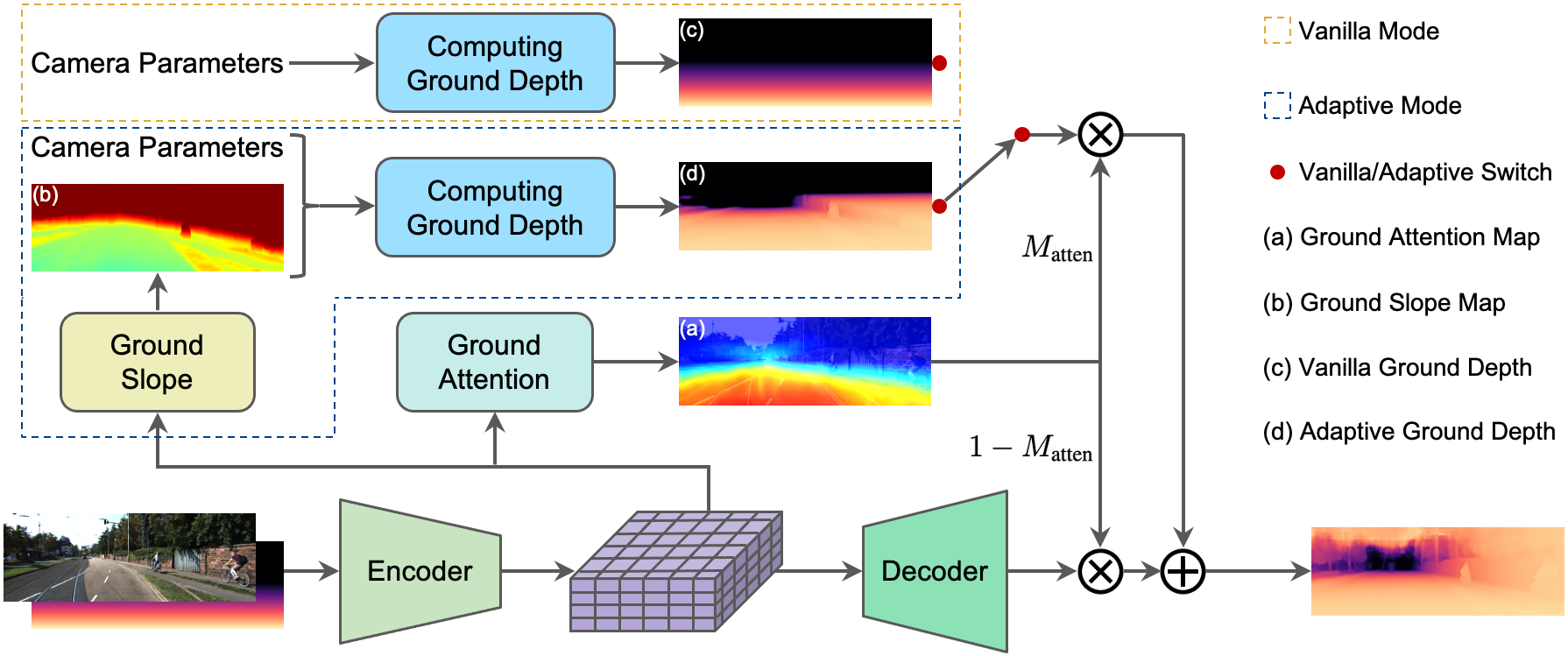}
 \caption{A schematic overview of the proposed ground embedding module integrated in a typical depth estimation network. Given camera parameters, we compute the ground depth and stack it with the input image to produce the ground depth-aware features through the encoder. A ground attention block is developed to selectively combine the ground depth and the residual depth (i.e., the direct output of the decoder) to form the final depth prediction. In the vanilla mode, the computation of planar ground depth involves camera parameters only, while in the adaptive mode, a ground slope block is additionally used to deal with the undulated ground.
 }
 \label{fig:workflow}
\end{figure*}

In light of the above observations, we develop a ground embedding module for monocular depth estimation, which we term \textbf{GEDepth}. It is a plug-in module that is lightweight and flexible to be incorporated into various depth estimation networks. As illustrated in Figure~\ref{fig:workflow}, given camera parameters, the module first computes ground depth, which is then fused with an input image to produce ground depth-aware features. Based on such features, the network generates residual depth and ground attention map, and the latter selectively combines the former with the ground depth to form final depth prediction. Although starting from planar ground to formulate ground embedding, our module is not constrained to this assumption, but is practically designed to be adaptive to handle ground undulation in real-world scenarios. Our approach decouples camera parameters from pictorial cues, thus improving the generalizability of depth estimation. This design also leads to an explicit utilization of ground, which reinforces the key information used for depth estimation as investigated in~\cite{dijk2019neural,hu2019visualization}: (1) the ground depth together with the ground attention map facilitates the learning of ground contact points; (2) the vanishing line (expressed in ground depth) coupled with the ground attention map locate the region around vanishing point readily. 

We summarize our main contributions as follows. (1) To our knowledge, this work provides the first plug and play module through ground embedding, which is able to assist various depth estimation networks in decoupling camera parameters from pictorial cues, and remarkably enhances their generalizability. (2) Our proposed module breaks the planar ground assumption and is capable of tackling ground undulation in realistic scenes. (3) Extensive experiments demonstrate that our approach compares favorably against the competing algorithms, and meanwhile, achieves more robust performance on a variety of cross-domain evaluations. Our code and model will be made available at \url{https://github.com/qcraftai/gedepth}.

\section{Related Work}
\label{sec:related_work}
\noindent\textbf{Monocular Depth Estimation.} 
As the pioneer work, Eigen et al. first adopt two networks to regress depth from coarse to fine in~\cite{eigen2014depth}, where a scale-invariant loss is developed to mitigate the scale ambiguity problem. 
A minimum reprojection loss together with an auto-masking loss~\cite{godard2019digging} are introduced to handle occlusions and ignore outlier pixels. 
Laina et al.~\cite{laina2016deeper} propose a reverse Huber loss by switching the penalty from $\ell_1$ to $\ell_2$ along with the increase of error. 
In~\cite{fu2018deep}, the depth regression objective is argued to be slow in convergence and easy to be trapped in a sub-optimal solution, instead, the task is modeled as an ordinal regression problem with a novel ordinal loss considering the order information between discrete depths. 
This method is further improved in~\cite{bhat2021adabins} via predicting the depth range of a scene into adaptive discrete bins with Transformers. Benefiting from the strong ability of learning long-range correlations, Transformers blocks have recently been incorporated either in the encoder~\cite{li2022depthformer} to model the relation of distant structures, or in the decoder~\cite{li2022binsformer} skipping the connection part~\cite{agarwal2023attention} to better fuse the features from multiple layers. 

However, these works primarily focus on designing more effective loss functions or exploiting more sophisticated architectures. Few of them has considered the impact of camera parameters in monocular depth estimation. GEDepth is orthogonal to the existing methods and can be applied as a plug-in module with these methods to further improve upon the current research achievements.    

\noindent\textbf{Geometric Priori.} 
3D geometry has been widely used in monocular depth estimation. For the unsupervised learning branch, the multi-view geometry is used to warp images from source view to target view to form a reconstruction loss to enforce consistency between views~\cite{godard2019digging, zhou2017unsupervised}. In addition to the photometric consistency,~\cite{hirose2021plg,mahjourian2018unsupervised} impose the geometric consistency between the point clouds generated from image depths with different camera poses. Another commonly used geometric priori is the normal constraint~\cite{long2021adaptive,qi2018geonet}, which enforces the consistency of normal vectors derived from the estimated and ground-truth depth. A local planar assumption is proposed in~\cite{lee2019big} using the multi-scale local planar guidance layers to construct a direct relation from internal features to predicted depth. Recent works~\cite{xing2022joint,yuan2021monocular} employ the planar parallax geometry, which divides the correspondence between consecutive images into planar homography and residual parallax, making the reconstruction more accurate and stable. 

Unlike the geometric cues applied in previous methods, we directly and explicitly make use of ground to provide the reference for more accurate and generalizable depth estimation. Beyond the ideal planar ground assumption, our approach is also designed to be able to adaptively process the undulated ground in realistic scenes. 

\section{Ground Embedding}
\label{section:Method}
\label{section:pe_module}
In this section, we start from introducing the formulation of ground depth given camera parameters. Based on this, we then present the two designs of ground embedding---the proposed plug-in module---from the ideal planar ground (vanilla) to the realistic undulated ground (adaptive).    

\subsection{Formulation of Ground Depth}
\label{sec:formulate}
Although monocular depth estimation is ill-posed by nature, for some specific areas, depth is deterministic given the camera parameters. In particular, if the height of a planar ground with respect to the world coordinate system is known (mostly can be read off from camera extrinsics), we can calculate the accurate depth of ground area in a closed-form solution. Unlike indoor scenes, there exists a ground region in nearly every image captured in self-driving scenes. This motivates us to take advantage of the ground depth as an accurate reference and a strong priori to alleviate the ambiguous problem, and eventually improve upon various depth estimation networks in autonomous driving.            

\begin{figure}[t]
\begin{center}
\includegraphics[width=\linewidth]{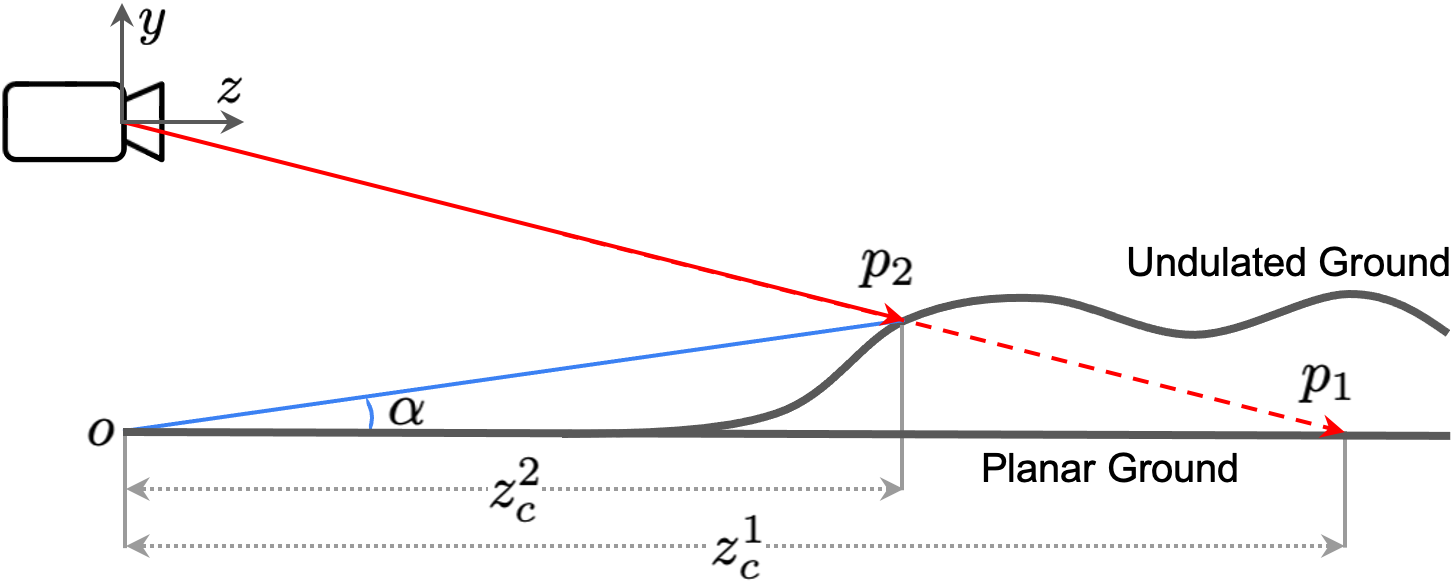}
\end{center}
   \caption{Illustration of different formulations of ground depth corresponding to one pixel in vanilla and adaptive modules. In the former, the depth $z_c^1$ is obtained by calculating the intersection point $p_1$ between the ray and planar ground. In the latter, the depth $z_c^2$ is derived based on the ground slope $\alpha$ formed by the projection of connection line $\overline{op}_2$ on the y-z plane and planar ground. 
   }
\label{fig:pe_module}
\end{figure}

\begin{figure}[t]
\begin{center}
\includegraphics[width=\linewidth]{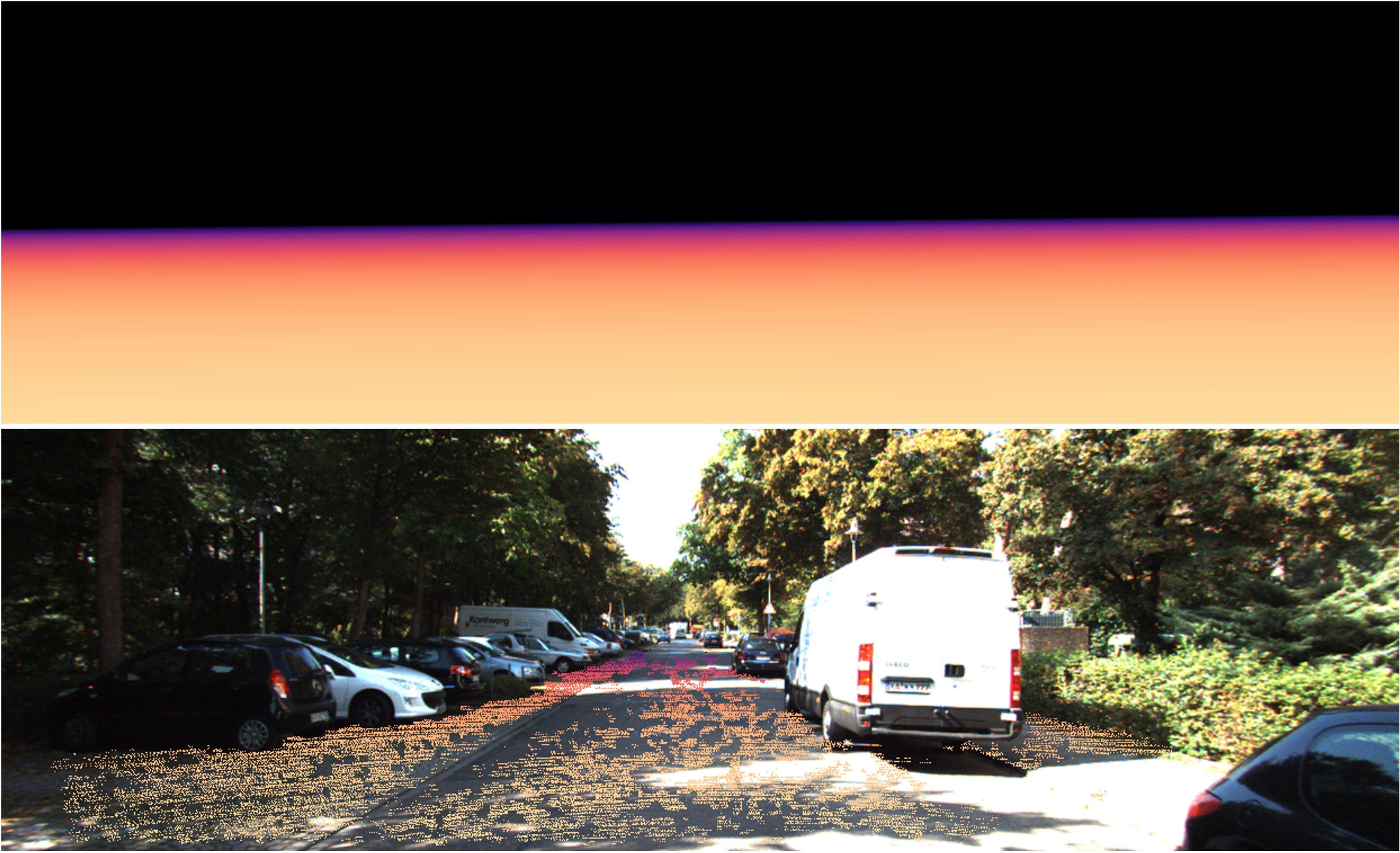}
\end{center}
   \caption{Top demonstrates the ground depth calculated from camera parameters (the dark region above the vanishing line indicates infinite depth). Bottom visualizes the pixels of which ground depth values are sufficiently close (i.e., relative error $<3\%$) to the sparse ground-truth values provided by LiDAR in KITTI. 
   }
\label{fig:pe_example}
\end{figure}

We denote the camera intrinsics as $\bm{K} \in \mathbb{R}^{3 \times 3}$, and the camera extrinsics as rotation matrix $\bm{R} \in \mathbb{R}^{3 \times 3}$ and translation vector $\bm{T} \in \mathbb{R}^{3 \times 1}$. Derived from the pinhole camera model, the transformation between a point $(x_w,y_w,z_w)$ in the world coordinate system and its projection $(u, v)$ in the pixel coordinate system can be described as:
\begin{equation}
z_c
\begin{bmatrix}
u \\
v \\
1 \\
\end{bmatrix}
=
\begin{bmatrix}
\bm{K}\;\;\;\bm{0}
\end{bmatrix}
\begin{bmatrix}
\bm{R}\;\;\;\bm{T} \\
\bm{0}^T\;\;1
\end{bmatrix}
\begin{bmatrix}
x_w \\
y_w \\
z_w \\
1
\end{bmatrix},
\label{eq.pinhole}
\end{equation}
where $z_c$ is the depth of pixel $(u, v)$ in the camera coordinate system. We then rewrite (\ref{eq.pinhole}) as: 
\begin{equation}
\begin{bmatrix}
x_w \\
y_w \\
z_w \\
\end{bmatrix}
=\bm{R}^{-1}
(\bm{K}^{-1}
\begin{bmatrix}
u \\
v \\
1 \\
\end{bmatrix}
z_c - \bm{T}). 
\label{eq.rewrite}
\end{equation}
We now represent the ray shooting from the camera optical center through each pixel as $r(u, v, z_c)$. For notation simplicity, we denote $\bm{R}^{-1}\bm{K}^{-1}$ as the matrix $\bm{A} = (a_{ij}) \in \mathbb{R}^{3 \times 3}$, and $\bm{R}^{-1}(-\bm{T})$ as the vector $\bm{B} = (b_i) \in \mathbb{R}^{3 \times 1}$. Thus, the parametric equation of the ray can be defined as: 
\begin{equation}
r(u, v, z_c) :
\begin{cases}
x_w = (a_{11}u +a_{12}v + a_{13}) z_c +  b_1  \\
y_w = (a_{21}u +a_{22}v + a_{23}) z_c +  b_2  \\
z_w = (a_{31}u +a_{32}v + a_{33}) z_c +  b_3  \\
\end{cases}
\label{eq.ray}
\end{equation}
Moreover, the planar ground with height $h$ can be described by a plane, which is determined by the point $(0, h, 0)$ in the plane and the normal vector $\overrightarrow{n}=(0,1,0)$:
\begin{equation}
y_w = h.
\label{eq.plane}
\end{equation}
As shown in Figure~\ref{fig:pe_module}, the ground depth can be formulated by calculating the depth of the intersection point between each ray and the plane. By combining (\ref{eq.ray}) and (\ref{eq.plane}), we obtain the ground depth of pixel $(u, v)$ as:
\begin{equation}
z_c = \\
\frac{h - b_{2}}{
a_{21}u + a_{22}v + a_{23}
}.
\label{eq.ge}
\end{equation}
Note we assign $z_c = 0$ to those pixels (i.e., vanishing line and above), through which the rays do not intersect with the plane. Figure~\ref{fig:pe_example} shows the ground depth examples in KITTI. 

\subsection{GEDepth-Vanilla}
\label{section:vanilla_pe}
As shown in Figure~\ref{fig:workflow}, the ground depth is first concatenated with a raw image to compose the input with ground depth priori. The encoder and decoder constitute the main body of the network, which can be instantiated by an existing depth estimation model with an architecture of either CNNs or Transformers. However, the ground depth priori is valid for the ground area only, but inaccurate for other non-ground regions. We therefore introduce the ground attention map $M_{\text{atten}}$, which is simply generated by a few convolutional layers based on the ground depth-aware features produced by the encoder. Each pixel in $M_{\text{atten}}$ represents the probability of that pixel belonging to the ground area. The ground attention map is then used to weighted combine the ground depth and its complementary counterpart that is the residual depth produced by the decoder to form the final depth prediction. 

\begin{figure*}[t]
 \begin{center}
\includegraphics[width=\linewidth]{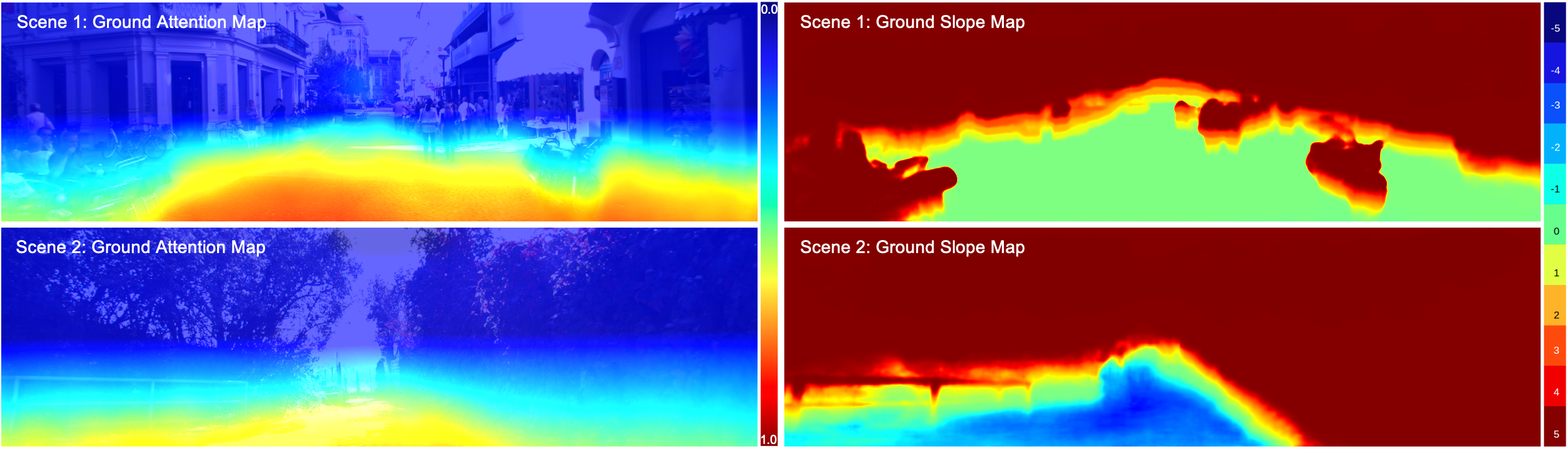}
\end{center}
 \caption{Visualization of the ground attention maps and the ground slope maps of two scenes in the test set of KITTI.
 }
\label{fig:attn_mask_vis_example}
\end{figure*}

It is noteworthy that the ground attention map is totally learned implicitly without using extra supervision provided from additional models such as ground segmentation. Instead, the whole network including the proposed module is trained by the original depth loss only, and it is found that the implicitly learned attention map can well separate the ground and other regions, as shown in Figure~\ref{fig:attn_mask_vis_example}.      

\subsection{GEDepth-Adaptive}
\label{section:dynamic_pe}
As aforementioned, the ground depth is valid for the planar ground area. However, in real-world driving scenarios, the undulated ground is not unusual, in particular in urban scenes where uphill and downhill roads are common, breaking the ideal planar ground assumption. To make the ground depth fulfill the real environment as closely as possible, we extend the proposed vanilla module to be adaptive to deal with the ground undulation. 

As illustrated in Figure~\ref{fig:pe_module}, we denote point $o$ as the projection of camera optical center to the ideal planar ground, and then connect $o$ with every surface point. For each image pixel $(u, v)$ corresponding to the surface point of undulated ground, we define its ground slope as the angle $\alpha$ between the ideal planar ground and the projection of connection line on the y-z plane. We can extend the ground description from the planar in (\ref{eq.plane}) to the undulated as:
\begin{equation}
y_w = \texttt{tan}(\alpha) z_c + h,
\label{eq.real_plane}
\end{equation}
where $z_c$ is the depth of pixel $(u, v)$ and $h$ is the height of planar ground, as defined in Section~\ref{sec:formulate}. Similarly, by combining (\ref{eq.ray}) and (\ref{eq.real_plane}), we get the undulated ground depth:
\begin{equation}
\begin{aligned}
z_c = \frac{b_2 - h}{\texttt{tan}({\alpha})-(a_{21}u +a_{22}v + a_{23})},
\label{eq.slope_ground}
\end{aligned}
\end{equation}
where only the ground slope value $\alpha$ is unknown. To obtain the value of $\alpha$, we introduce the ground slope map $M_{\text{slope}}$, 
which is also generated based on the ground depth-aware features, as shown in Figure~\ref{fig:workflow}. Here we pre-define a set of $N$ discrete angles: $\{\tau_i \in [-\frac{\pi}{6}, \frac{\pi}{6}] \mid i=1,...,N\}$ in accordance with the slope statistics in training data. Each pixel in the map represents the approximated ground slope, which is computed by softly combining the pre-defined angles with the predicted probability distribution $\{p_i \in [0, 1] \mid \sum_{i = 1}^N p_i = 1\}$ over the $N$ classes or angles:
\begin{equation}
\begin{aligned}
\Hat{\alpha} = \sum_{i=1}^N p_i \tau_i. 
\label{eq.softmax}
\end{aligned}
\end{equation}
\hspace{-1mm}We finally substitute $\alpha$ in (\ref{eq.slope_ground}) with its estimated $\Hat{\alpha}$ in (\ref{eq.softmax}) to calculate the undulated ground depth. In practice, by leveraging the available ground-truth depth of sparse ground surface points, we can transform (\ref{eq.slope_ground}) to the following equation to provide the sparse supervision of ground slope in order to facilitate the ground slope map learning:
\begin{equation}
\begin{aligned}
\alpha = \texttt{arctan}\left(\frac{b_2-h}{z_c}+a_{21}u +a_{22}v + a_{23}\right).
\label{eq.slope_alpha}
\end{aligned}
\end{equation}

\subsection{Optimization}
GEDepth is a simple plug and play module that can be integrated into an existing depth estimation network with minimum change of the original model. We train the whole network to optimize the following total objective:
\begin{equation}
\begin{aligned}
L_{\text{total}} = L_{\text{reg}}(z,\hat{z}) + \mathbbm{1} \lambda_{\text{cls}} L_{\text{cls}}(\alpha,\hat{\alpha}),
\label{eq.total_loss}
\end{aligned}
\end{equation}
where $L_{\text{reg}}$ is the depth regression loss used in the original network and $L_{\text{cls}}$ is the classification loss in ground slope learning, $z$ and $\hat{z}$ denote the ground-truth and predicted depth, and the ground slope $\alpha$ and $\hat{\alpha}$ are defined similarly. $\lambda_{\text{cls}}$ is the classification loss weight, and $\mathbbm{1}$ indicates if the applied module is GEDepth-Adaptive (in other words, the total objective involves the original regression loss only for GEDepth-Vanilla).

\section{Experiments}
\label{section:experiments}
In this section, we first introduce the experimental setup including datasets, metrics and implementation details. We then report extensive comparisons with the state-of-the-art methods on the popular benchmarks. A variety of generalization and ablation studies are finally conducted for the in-depth understanding of the proposed approach. 

\subsection{Experimental Setup}
\label{section:setup}
\noindent\textbf{Datasets.} 
We extensively evaluate our approach on the two depth estimation datasets: KITTI~\cite{geiger2013vision} and Dense Depth for Autonomous Driving (DDAD)~\cite{guizilini20203d}. KITTI is used as the de facto benchmark for depth evaluation. We follow the standard Eigen split~\cite{eigen2014depth} that consists of 23,158 images for training and 697 for test. We use the crop defined in~\cite{garg2016unsupervised} and upsample predicted depth to the ground-truth resolution for evaluation. DDAD is a more challenging benchmark with rich geographic diversity, multiple camera views, and long prediction range. It contains 150 scenes (12,650 images per camera) for training and 50 scenes (3,950 images per camera) for test. We follow the same protocol as defined in~\cite{guizilini2021sparse} to use the four camera views including forward, backward, left forward, and right forward. 

\noindent\textbf{Metrics.} 
We follow previous works and adopt a series of evaluation metrics widely used in the community, including absolute relative distance (Abs Rel), squared relative distance (Sq Rel), root mean squared error (RMSE), root mean squared error in log space (RMSE-log), and scale-invariant logarithmic error (SILog). However, it is argued in~\cite{pinar20222d} that the traditional depth estimation metrics mainly consider 2D global pixel-wise error but lack of 3D structural awareness. Therefore, we additionally report a set of metrics recently proposed in~\cite{pinar20222d} that are well suited to evaluate 3D geometry, including 3D point cloud F-score (F-score) and 3D point cloud intersection over union (IoU).  

\noindent\textbf{Implementation Details.} In order to thoroughly evaluate GEDepth, we plug the module into four representative networks including DepthFormer~\cite{li2022depthformer}, PixelFormer~\cite{agarwal2023attention}, BinsFormer~\cite{li2022binsformer} and BTS~\cite{lee2019big}, which represent the state-of-the-art depth estimation models in both Transformers and CNNs. We implement our approach based on their open-sourced codebases. We train each network on 8 NVIDIA V100 GPUs and follow the original training configuration (such as batch size, learning rate, training epochs, optimizer, etc.) of each corresponding algorithm (see more details in the supplementary material). We set $\lambda_{\text{cls}} = 0.1$ in (\ref{eq.total_loss}), and $h = 1.65$ and $0$ for KITTI and DDAD respectively according to their defined ground heights. In line with the ground slope distribution in training data, we use 11 discrete angles evenly distributed in $[-5, 5]$.

\subsection{Comparison with State-of-the-Art Results}

\begin{table*}[t]
\small
\begin{center}
\begin{tabular}{lccccccc}
\hline
Method & Abs Rel $\downarrow$ &  Sq Rel $\downarrow$ & RMSE $\downarrow$ & RMSE-log $\downarrow$ & SILog $\downarrow$ & F-score $\uparrow$ &  IoU $\uparrow$\\
\hline
Eigen~\cite{eigen2014depth} & 0.190 & 1.515 & 7.156 & 0.270 & -&-&-\\
DORN~\cite{fu2018deep}         & 0.072 & 0.307 & 2.727 & 0.120 &  -&-&-\\
DPT~\cite{ranftl2021vision}          & 0.062 & 0.222 & 2.575 & 0.092 &-&-&-\\
AdaBins~\cite{bhat2021adabins}       & 0.058 & 0.190 & 2.360 & 0.088 &-&-&-\\
NeW CRFs~\cite{yuan2022new}     & 0.052 & 0.155 & 2.129 & 0.079 &-&-&-\\
\hline
BTS~\cite{lee2019big}& 0.060 & 0.213 & 2.537 & 0.094 & 8.368 & 0.472 & 0.318\\
BTS (GE-Vanilla)& 0.058 & 0.196 & 2.420 & 0.090 & 8.170 & 0.485 & 0.329\\
BTS (GE-Adaptive)& \textbf{0.056} & \textbf{0.193} & \textbf{2.401} & \textbf{0.089} & \textbf{8.087} & \textbf{0.487} & \textbf{0.331}\\ \hline
DepthFormer~\cite{li2022depthformer}  & 0.052 & 0.156 & 2.133 & 0.079 & 7.210& 0.493 & 0.336\\
DepthFormer (GE-Vanilla)& 0.049 & 0.144 & 2.063 & 0.077 & 6.983 & 0.507 & 0.349\\
DepthFormer (GE-Adaptive)& \textbf{0.048} & \textbf{0.142} & \textbf{2.050} & \textbf{0.076} & \textbf{6.982} &\textbf{0.515}&\textbf{0.356}\\ 
\hline
PixelFormer~\cite{agarwal2023attention}  & 0.051 & 0.149 & 2.081 & 0.077 &7.061& 0.496 &0.340\\
PixelFormer (GE-Vanilla)& 0.050 & 0.145 & 2.071 & 0.077  & 7.059 & 0.496 & 0.340\\
PixelFormer (GE-Adaptive)& \textbf{0.049} & \textbf{0.143} & \textbf{2.054} & \textbf{0.076} &\textbf{6.991}&\textbf{0.507}&\textbf{0.349}\\ \hline
BinsFormer~\cite{li2022binsformer}  & 0.052 & 0.151 & 2.098 & 0.079 &7.266& 0.488 &0.333\\
BinsFormer (GE-Vanilla)& 0.051 & 0.146 & 2.080 & 0.078 &7.101&0.491&0.335\\
BinsFormer (GE-Adaptive)& \textbf{0.050} & \textbf{0.143} & \textbf{2.052} & \textbf{0.077} &\textbf{7.084}&\textbf{0.502}&\textbf{0.344}\\ \hline
\end{tabular}
\end{center}
\vspace{-4pt}
\caption{Comparison of GEDepth (in both vanilla and adaptive modes) and the state-of-the-art methods on KITTI. Groups 2-5 correspond to the four representative methods integrated with the proposed ground embedding modules.} 
\vspace{-2pt}
\label{tab:kitti_results}
\end{table*}

\begin{figure*}[!htb]
\centering
\includegraphics[width=1.0\linewidth]{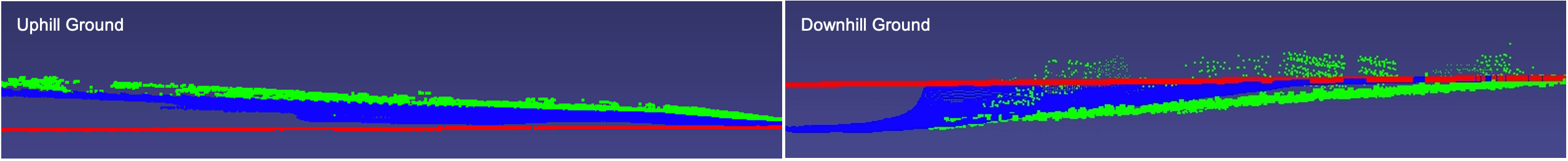}
 \caption{
Visualization of the side view of the ground depth unprojected to 3D in the uphill and downhill scenarios in KITTI. In each scene, the red and blue points respectively denote the ground depth computed by GEDepth-Vanilla and GEDepth-Adaptive, and the green points correspond to the ground-truth point clouds scanned by LiDAR. 
 }
 \label{fig:dynamic_slope_vis}
\end{figure*}

\noindent\textbf{KITTI.} We extensively compare our approach with a variety of the state-of-the-art methods in Table~\ref{tab:kitti_results}. As we can see, the performance of leading algorithms tends to saturate, e.g., Abs Rel of several methods has reached a plateau around 0.052. By equipped with GEDepth, the results of a broad range of representative methods are clearly improved, indicating that our approach is applicable to various depth estimation networks. In addition, the improvement by adaptive module is overall more significant than that of vanilla module, which validates our design of modeling ground undulation. We further show the ground attention maps in Figure~\ref{fig:attn_mask_vis_example}. Although the maps are implicitly learned without any direct supervision, they segregate the ground regions reasonably well, thus providing a solid foundation for the combination of ground depth and residual depth. In this figure, we also visualize the paired ground slope maps, which deliver the plausible descriptions of ground undulation. As demonstrated in Figure~\ref{fig:dynamic_slope_vis}, we unproject the ground depth calculated by vanilla and adaptive modules to 3D, where the latter is found to be more effective in fitting the real ground points (captured by LiDAR). This evidently exhibits the effectiveness of the proposed undulated ground modeling in the adaptive module.

\noindent\textbf{DDAD.} 
We also evaluate on this multi-view dataset where four camera views are simultaneously considered. As compared in Table~\ref{tab:ddad_results}, GEDepth is superior to each counterpart method and again achieves consistent improvement over all metrics. This verifies that our approach is applicable to not only different networks but also different datasets. Furthermore, larger improvement is observed on DDAD than KITTI, which shows the strong potential of our module in tackling more challenging scenarios. 

\begin{table*}
\small
\begin{center}
\begin{tabular}{lcccccccc}
\hline
Method &Abs Rel $\downarrow$ &  Sq Rel $\downarrow$ & RMSE $\downarrow$ & RMSE-log $\downarrow$ & SILog $\downarrow$ &  F-score $\uparrow$ &IoU $\uparrow$ \\
\hline
PackNet-SAN~\cite{guizilini2021sparse} & 0.187 & 2.776 & 11.936 & 0.276 & - & - & -\\
\hline
BTS~\cite{lee2019big} & 0.162 & 2.492 & 11.466 & 0.259 & 24.314 & 0.636 & 0.478\\
BTS (GE-Vanilla) & 0.158 & 2.377 & 11.219 & \textbf{0.253} & 23.863 & 0.636 & 0.479\\

BTS (GE-Adaptive) & \textbf{0.156} & \textbf{2.360} & \textbf{11.186} & \textbf{0.253} & \textbf{23.761} & \textbf{0.642} & \textbf{0.485}\\
\hline
DepthFormer~\cite{li2022depthformer} & 0.152 & 2.230 & 11.051 & 0.246 &22.629 & 0.649 & 0.493\\
DepthFormer (GE-Vanilla) & 0.149 & 2.121 & 10.790 & 0.240 & 22.437 & 0.650 & 0.493\\
DepthFormer (GE-Adaptive) & \textbf{0.145} & \textbf{2.119} & \textbf{10.596} & \textbf{0.237} & \textbf{22.190} & \textbf{0.656} & \textbf{0.500}\\
\hline
PixelFormer~\cite{agarwal2023attention} & 0.151 & 2.140 & 10.920 & 0.242 & 22.311 & 0.659 & 0.502\\
PixelFormer (GE–Vanilla) & 0.148 & 2.123 & 10.848 & \textbf{0.241} & 22.272 & 0.660 & 0.504\\
PixelFormer (GE-Adaptive) & \textbf{0.145} & \textbf{2.122} & \textbf{10.803} & \textbf{0.241} & \textbf{22.268} & \textbf{0.661} & \textbf{0.505}\\
\hline
BinsFormer~\cite{li2022binsformer} & 0.149 & 2.142 & 10.866 & 0.244 & 22.513 & 0.653 & 0.496\\
BinsFormer (GE-Vanilla) & 0.146 & 2.109 & 10.561 & \textbf{0.235} & 22.252 & 0.658 & 0.502\\
BinsFormer (GE-Adaptive) & \textbf{0.145} & \textbf{2.101} & \textbf{10.459} & \textbf{0.235} & \textbf{22.060} & \textbf{0.659} & \textbf{0.504}\\
\hline
\end{tabular}
\end{center}
\caption{Comparison of GEDepth (in both vanilla and adaptive modes) and the state-of-the-art methods on DDAD. Groups 2-5 correspond to the four representative methods integrated with the proposed ground embedding modules.}
\label{tab:ddad_results}
\end{table*}

\subsection{Generalization Study} 
\label{section:generalization}
\begin{figure}[t]
		\centering
            \includegraphics[width=1.0\linewidth]{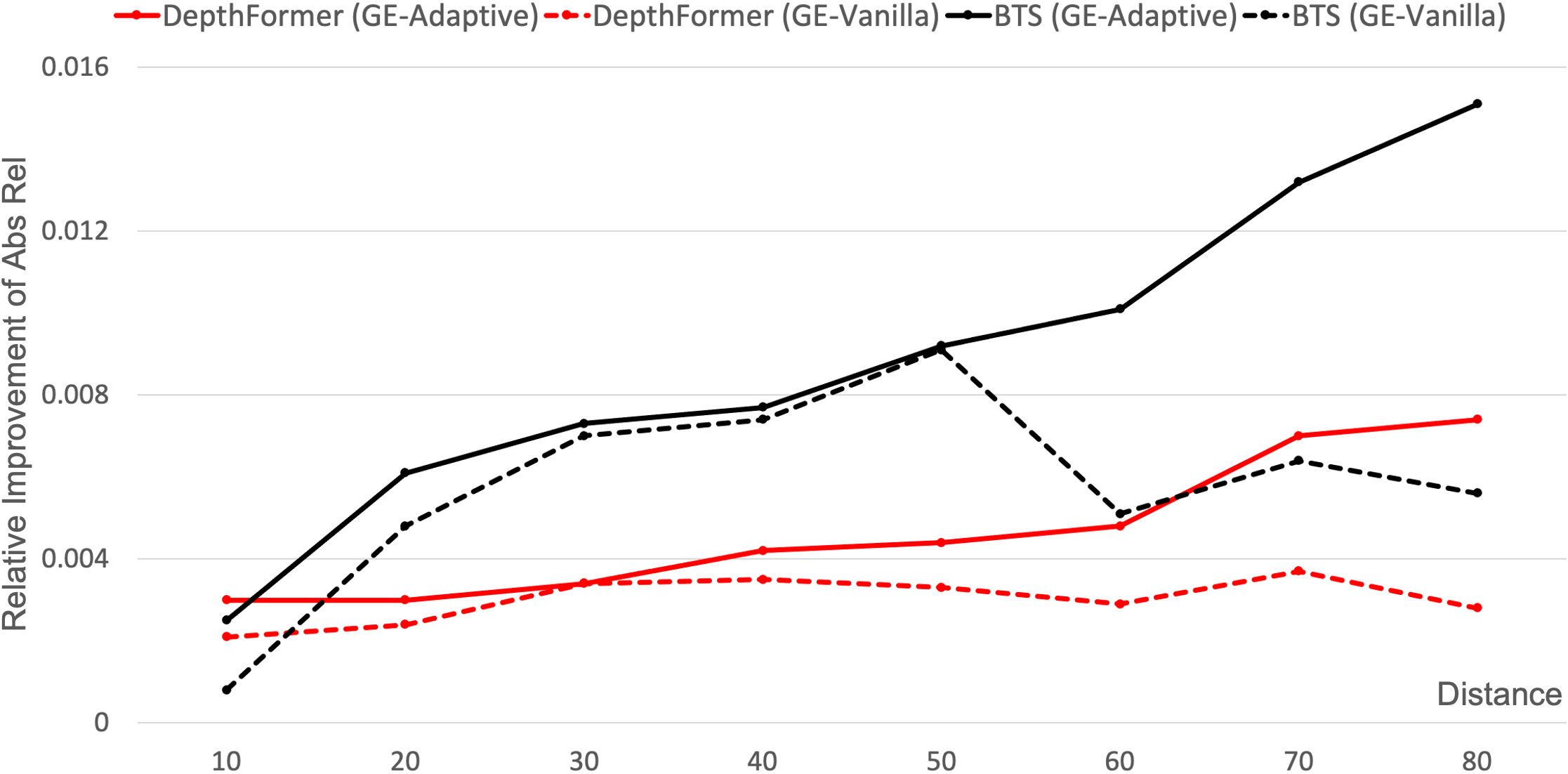}
\caption{
Comparison of the relative performance improvement over DepthFormer and BTS by GEDepth-Vanilla and GEDepth-Adaptive at different distance intervals on KITTI.
}
\label{fig:kitti_distance}
\end{figure}

\noindent\textbf{Generalization on Distances.} 
We first evaluate the generalization effect of depth estimation by our approach in different distances. We use DepthFormer and BTS as two representative methods based on Transformers and CNNs, respectively. Figure~\ref{fig:kitti_distance} compares the relative improvements over the two methods under different distance intervals on KITTI. It is observed that as the distance increases, GEDepth-Vanilla and GEDepth-Adaptive both provide notable performance gains. As for the farther distances (e.g., $>60$m), the improvement by adaptive module is more significant due to a higher chance to encounter ground undulation in the distant areas. This comparison indicates that our approach improves the generalization in the whole distances, especially for the long range.               

\begin{figure}[t]
\centering
\includegraphics[width=1.0\linewidth]{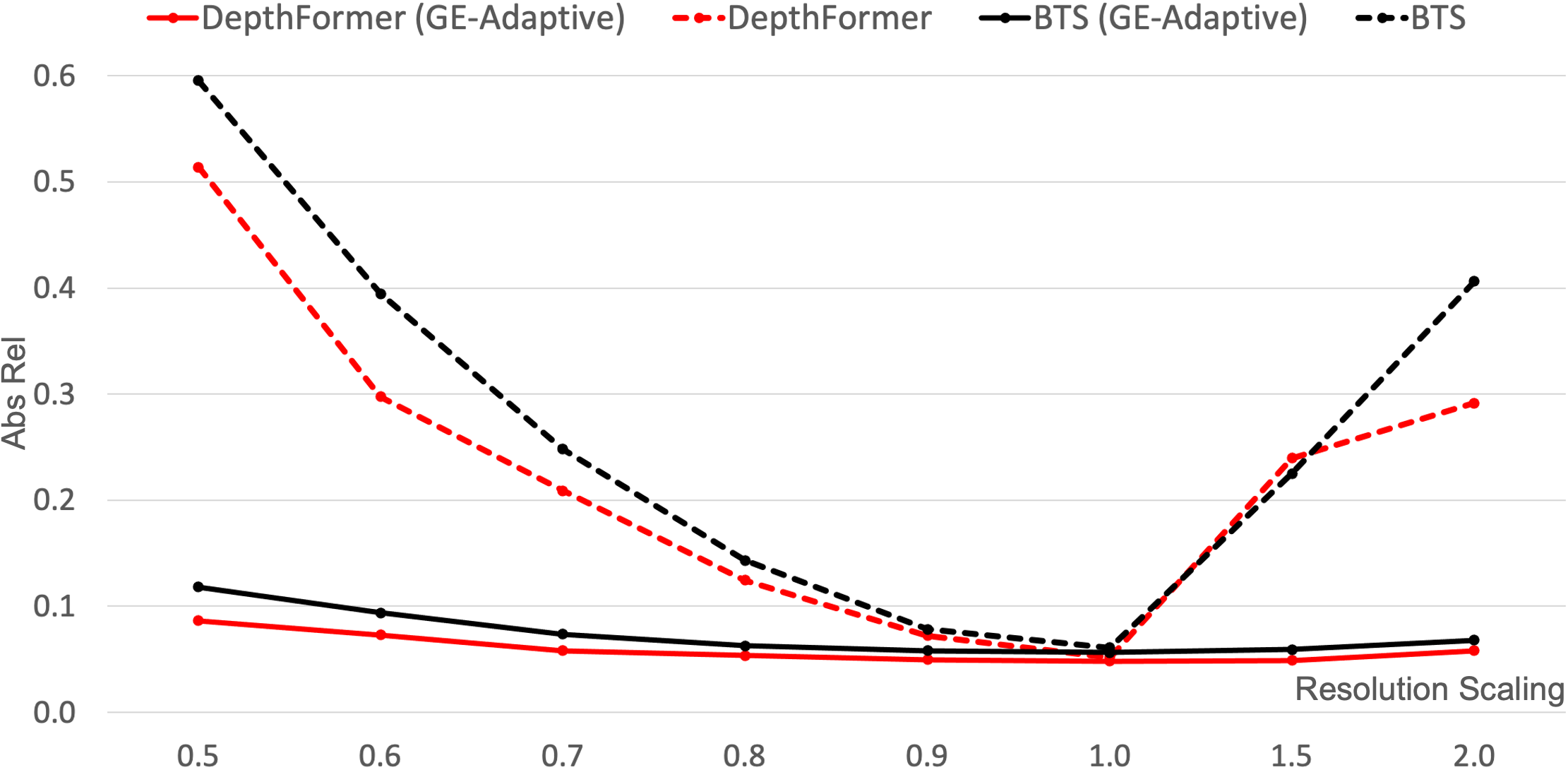}
\caption{
Comparison of the impact of image resolution change to DepthFormer and BTS on KITTI. GEDepth-Adaptive enables the two methods to be significantly more robust to the change.  
}
\label{fig:kitti_resize}
\end{figure}

\noindent\textbf{Generalization on Resolutions.} 
Here we study how the proposed ground embedding module impacts the generalization of depth estimation with respect to different input image resolutions (i.e., change of intrinsics). As compared in Figure~\ref{fig:kitti_resize}, DepthFormer and BTS are both severely sensitive to the change in image resolution. In the contrary, GEDepth-Adaptive makes the two methods substantially more resistant to the resolution change. This clearly demonstrates the efficacy of our approach in decoupling camera intrinsics as well as the resulting benefit in mitigating the scale ambiguity problem that is inherent for the original depth estimation networks. 

\begin{figure}[t]
\centering
\includegraphics[width=1.0\linewidth]{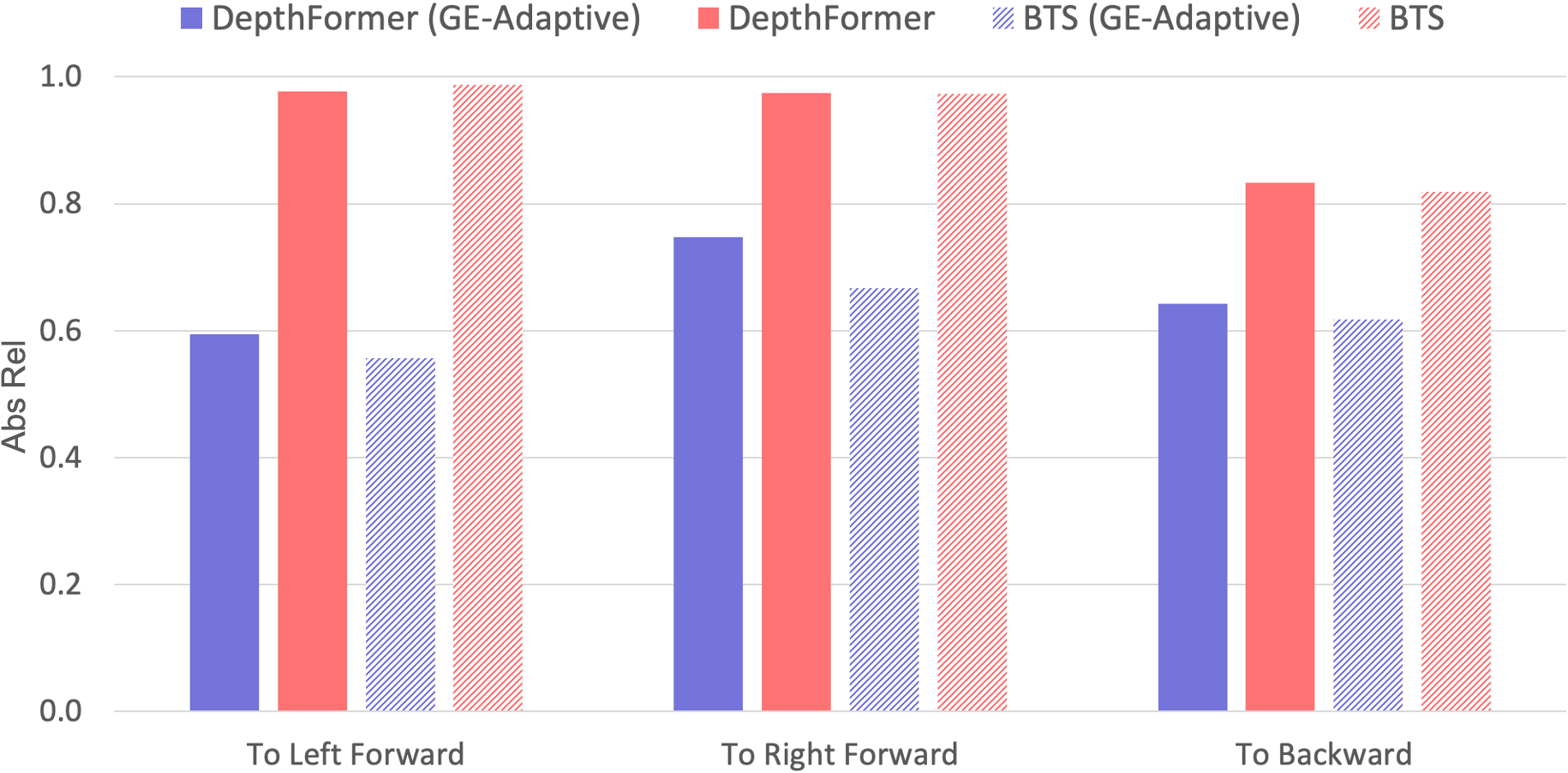}
\caption{
Comparison of the view-transferring influence (trained on the forward view and then tested on the left forward, right forward and backward views) for DepthFormer and BTS on DDAD. GEDepth-Adaptive significantly improves the generalizability on cross-camera views of the two methods.
}
\label{fig:ddad_train_cam1}
\end{figure}

\noindent\textbf{Generalization on Camera Views.} 
Next we investigate the enhancement of our proposed module to depth estimation on different camera views (i.e., change of both intrinsics and extrinsics). We perform the experiment on the multi-view dataset DDAD, where a network is trained using the forward view only and then evaluated on the other three views. As we can see in Figure~\ref{fig:ddad_train_cam1}, DepthFormer and BTS are both tremendously degraded when they are transferred from the forward to other views. As a contrast, GEDepth-Adaptive largely improves their view-transferring performance. This cross-camera test again validates the advantage of decoupling camera parameters from pictorial cues by our module in promoting the generalization capability.     

\noindent\textbf{Generalization on Datasets.} 
At last, we probe into the generalization of our approach across datasets (i.e., change of both intrinsics and extrinsics as well as data distribution). As shown in Table~\ref{tab:cross_dataset}, we first train the network on one dataset and then evaluate on the other (without fine-tuning). Compared with DepthFormer, GEDepth-Adaptive improves the results by a large margin on either way, revealing that our ground embedding is also more robust to the change of data distribution in addition to the change of camera parameters, since the calculated ground depth is able to provide vital reference in both datasets.  

\subsection{Ablation Study}
\label{section:ablation_study}

\noindent\textbf{Improvement on Different Regions.}  
To better understand the contribution of our approach, we take advantage of the ground attention map to partition an image into ground and non-ground regions, then evaluate the performance of two regions separately. Table~\ref{tab:diff_area} shows that the overall improvement attributes to not only ground but also non-ground region, and the improvement for the latter is even more notable, indicating the merit of explicit modeling of ground to facilitate depth estimation for non-ground region.

\begin{table}
\small 
\begin{center}
\begin{tabular}{lcccc}
\hline
Dataset & Method &Abs Rel $\downarrow$ & RMSE $\downarrow$ \\
\hline
\multirow{2}*{K $\rightarrow$ D} & DepthFormer & 0.644 & 17.083\\
~ & GE-Adaptive & \textbf{0.261} & \textbf{16.132}\\
\hline

\multirow{2}*{D $\rightarrow$ K} & DepthFormer & 0.149 & 4.132\\
~ & GE-Adaptive & \textbf{0.104} & \textbf{3.398}\\
\hline

\end{tabular}
\end{center}
\caption{Comparison of the cross-dataset performance based on DepthFormer. K $\rightarrow$ D denotes the networks trained on KITTI and tested on DDAD, and vice versa. GEDepth-Adaptive brings significant generalization enhancement to the method. 
}
\label{tab:cross_dataset}
\end{table}

\begin{table}
\small
\begin{center}
\begin{tabular}{lcccc}
\hline
Region & Method &Abs Rel $\downarrow$ & RMSE $\downarrow$ \\
\hline
\multirow{2}*{Ground} & DepthFormer & 0.032 & 0.431\\
~ & GE-Adaptive & \textbf{0.028} & \textbf{0.399}\\
\hline
\multirow{2}*{Non-Ground} & DepthFormer & 0.078 & 3.225\\
~ & GE-Adaptive & \textbf{0.074} & \textbf{3.104}\\
\hline
\end{tabular}
\end{center}
\caption{Comparison of the ground and non-ground performance based on DepthFormer. GEDepth-Adaptive improves the method on both regions, while the non-ground improvement is greater.}
\label{tab:diff_area}
\end{table}

\begin{figure}[t]
\centering
\includegraphics[width=1.0\linewidth]{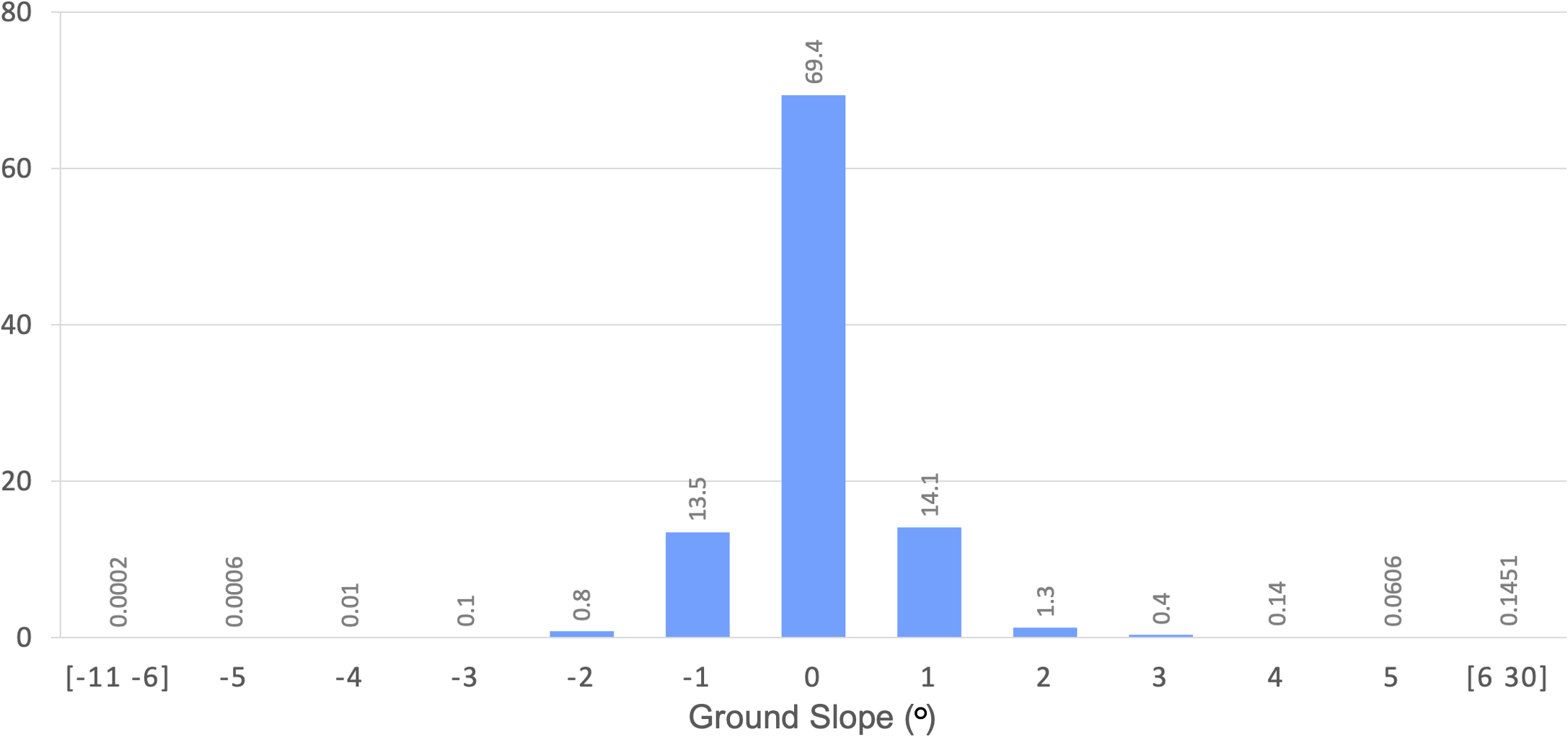}
\caption{
Distribution of ground slope in training data of KITTI.
}
\label{fig:slope_dist}
\end{figure}

\noindent\textbf{Ground Slope.}  
We also study the influence of range and binning for ground slope in GEDepth-Adaptive. As shown in Figure~\ref{fig:slope_dist}, according to the statistics in training data, the majority of ground slop is within $[-5, 5]$, and very few is beyond this range. Table~\ref{tab:kitti_slope_ablation} shows that GEDepth-Adaptive is robust to the range and binning in a wide scope. However, reducing the binning or increasing the range does not bring further improvement. This is because the network cannot differentiate very subtle slope difference simply from an image, and the samples of steep ground are too rare.    

\noindent\textbf{Vanilla and Adaptive.} 
Although the two types of modules have been compared in Tables \ref{tab:kitti_results} and \ref{tab:ddad_results}, the overall performance on a whole dataset may not fully convey the distinction between them as the planar ground dominates like shown in Figure~\ref{fig:slope_dist}. To gain a better understanding, we perform a further experiment on a subset consisting of obvious sloping scenes from KITTI. As compared in Table~\ref{tab:eavl_slope_sample}, the improvement difference between GEDepth-Vanilla and GEDepth-Adaptive is considerably amplified, suggesting the promising potential of the adaptive module in tackling more challenging scenes with undulated ground.  

\noindent\textbf{Inference Latency.}  
We in the end take a closer look into the impact of our approach to network inference latency, which is an equally important property for a lightweight plug-in module. As can be seen in Table~\ref{tab:inference_time}, GEDepth remarkably improves the performance, but with a cost of negligible increase in the network parameters and inference latency.

\begin{table}
\small
\begin{center}
\begin{tabular}{lccc}
\hline
Range & Binning & Abs Rel $\downarrow$ & RMSE $\downarrow$ \\
\hline
-5 to 5 & $0.1$ & 0.049 & 2.087\\
\hline
-5 to 5 & $0.5$ & 0.049 & 2.080\\
\hline
-5 to 5 & $1.0$ & 0.048 & 2.050\\
\hline
-5 to 5 & $2.0$ & 0.049 & 2.063\\
\hline
-5 to 10 & $1.0$ & 0.049 & 2.076\\
\hline
-5 to 20 & $1.0$ & 0.049 & 2.068\\
\hline
-5 to 30 & $1.0$ & 0.049 & 2.080\\
\hline
\end{tabular}
\end{center}
\caption{Comparison of GEDepth-Adaptive (with DepthFormer) using different ranges and binnings for ground slope on KITTI.}
\label{tab:kitti_slope_ablation}
\end{table}

\begin{table}
\centering
\resizebox{\columnwidth}{!}{ 
\begin{tabular}{lcccc}
\hline
Method &Abs Rel $\downarrow$ &  Sq Rel $\downarrow$ & RMSE $\downarrow$ & RMSE-log $\downarrow$ \\
\hline
DepthFormer & 0.067 & 0.180 & 2.163 & 0.089 \\
GE-Vanilla & 0.064 & 0.167 & 2.140 & 0.086 \\
GE-Adaptive & \textbf{0.057} & \textbf{0.156} & \textbf{2.100} & \textbf{0.080} \\
\hline
\end{tabular}
}
\vspace{3pt}
\caption{Comparison of GEDepth-Vanilla and GEDepth-Adaptive (with DepthFormer) on a subset that is made up of obvious sloping scenes selected from KITTI.
}
\label{tab:eavl_slope_sample}
\end{table}

\begin{table}
\small
\begin{center}
\begin{tabular}{lcccc}
\hline
Method & Params & Latency $\downarrow$ & Abs Rel $\downarrow$ & RMSE $\downarrow$ \\
\hline
DepthFormer & 274M & 179ms & 0.052 & 2.143\\
\hline
GE-Vanilla & 275M & 182ms & 0.049 & 2.063\\
\hline
GE-Adaptive & 277M & 185ms & 0.048 & 2.050\\
\hline
\end{tabular}
\end{center}
\caption{Comparison of network parameters, inference latency, and estimation accuracy for DepthFormer using GEDepth on KITTI.
}
\label{tab:inference_time}
\end{table}

\section{Conclusion}
\label{section:conclusion}
We have presented a simple module GEDepth based on the proposed ground embedding to decouple camera parameters and pictorial cues to enhance generalizability for monocular depth estimation. Beyond the planar ground, our approach is capable of handling undulated ground. It is a plug and play module that can be integrated into various depth estimation networks requiring little effort. GEDepth achieves the state-of-the-art results on both KITTI and DDAD, and meanwhile, provides substantial improvement on the comprehensive cross-domain evaluations.    

{\small
\bibliographystyle{ieee_fullname}
\bibliography{egbib}
}

\appendix
\section*{Appendix}
In Section~\ref{config}, we describe the detailed training configurations. Section~\ref{where} provides more ablation study on embedding ground depth. Section~\ref{lack} presents the performance in the images that are lack of ground. Section~\ref{hyper} studies the hyper-parameter. Section~\ref{viz} shows qualitative results. 

\section{Training Configurations}
\label{config}
In the paper, we have extensively evaluated GEDepth with the following four representative networks, 
which represent the state-of-the-art methods of monocular depth estimation in both Transformers and CNNs. For fair comparisons, we follow the original training configuration of each method when training GEDepth.

\begin{itemize}[noitemsep,topsep=0pt]
\item \textbf{DepthFormer} and \textbf{PixelFormer:} We set the bath size as 16 on 8 GPUs and the initial learning rate as 1e-4. We use the cosine annealing learning rate for 38.4K iterations and apply the linear learning rate warm-up strategy for the first 30\% iterations. 
\item \textbf{BinsFormer:} We set the batch size to 16 on 8 GPUs and the initial learning rate to 1e-4. We adopt the one-cycle learning rate for 38.4K iterations and use the linear learning rate warm-up for the first 30\% iterations. 
\item \textbf{BTS:} We set the batch size as 64 on 8 GPUs and the initial learning rate as 1e-4. We utilize the cosine annealing learning rate scheduler for 24 epochs without using the warm-up strategy.
\end{itemize}

\noindent AdamW is used as the optimizer for all networks above. 

\section{Where to Embed Ground Depth}
\label{where}
As illustrated in Figure 2, our approach originally embeds the ground depth in the input. Here we evaluate embedding the ground depth in the encoder as an alternative. Table~\ref{tab:where} shows that embedding in the encoder also improves over the baseline, but is inferior to the original embedding. 

\begin{table}[h]
\small
\begin{center}
\begin{tabular}{lcccc}
\hline
Method & Abs Rel $\downarrow$ & RMSE $\downarrow$ & SILog $\downarrow$\\
\hline
DepthFormer & 0.052 & 2.133 & 7.210\\
\hline
GE-Vanilla (encoder) & 0.050 & 2.071 & 7.074\\
\hline
GE-Vanilla (input) & 0.049 & 2.063 & 6.983\\
\hline
GE-Adaptive (encoder) & 0.049 & 2.070 & 7.072\\
\hline
GE-Adaptive (input) & 0.048 & 2.050 & 6.982\\
\hline
\end{tabular}
\end{center}
\vspace{-1mm}
\caption{Comparison of where to embed ground depth in GEDepth.}
\label{tab:where}
\end{table}

\section{Scenes Lacking of Ground}
\label{lack}
\begin{figure}
    \centering
    \includegraphics[width=\linewidth]{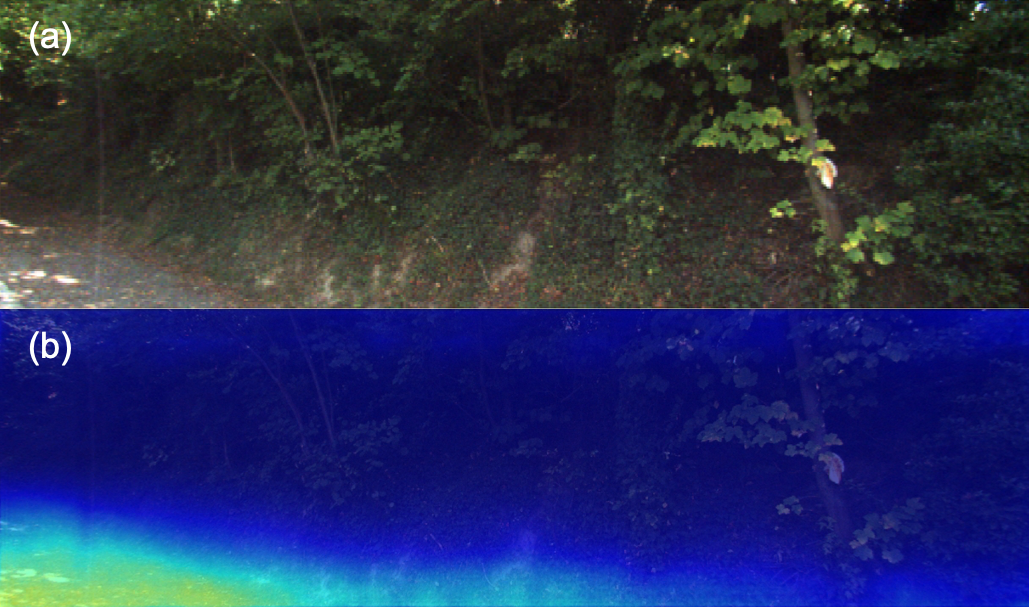}
    \caption{Visualization of the predicted ground attention map (b) in the scene (a) where ground is barely present.}  
    \label{fig:lack} 
\end{figure}

Although ground is ubiquitous in the camera images captured by autonomous driving vehicles, here we probe into how GEDepth performs in the rare case where ground is barely present in the test set of KITTI. As shown in Figure~\ref{fig:lack}, our approach is still able to predict reasonably accurate ground attention map. Table~\ref{tab:lack} reports that the overall result in this scene degrades compared to the common ones where ground is apparently present, while our approach still improves over the
baseline.   

\begin{table}
\small
\begin{center}
\begin{tabular}{lcccc}
\hline
Method &    Abs Rel $\downarrow$ & RMSE $\downarrow$   \\
\hline
DepthFormer & 0.140 & 1.018 \\
\hline
GE-Vanilla  & 0.073 & 0.543  \\
\hline
GE-Adaptive & \textbf{0.068} & \textbf{0.502} \\
\hline
\end{tabular}
\end{center}
\vspace{-1mm}
\caption{Comparison of baseline (DepthFormer) with GEDepth-Vanilla and GEDepth-Adaptive in the scenario where ground is barely present as shown in Figure~\ref{fig:lack}.}
\label{tab:lack}
\end{table}

\section{Hyper-Parameter}
\label{hyper}
We next evaluate how our approach behaves with different values of $\lambda_{\text{cls}}$, the classification loss weight for ground slope learning in Equation (10) of the paper. As shown in Table~\ref{tab:lambda}, our approach is overall robust to the values of this hyper-parameter in a reasonable range ($\lambda_{\text{cls}} = 0.10$ is the default value used in our experiments). 

\begin{table}[h]
  \small
  \centering
  \begin{tabular}{@{}lc@{}lc@{}lc@{}lc@{}lc@{}lc@{}lc@{}lc@{}lc@{}lc@{}lc@{}}
    \toprule
    $\lambda_{\text{cls}}$  &Abs Rel& &Sq Rel & &RMSE & & RMSE-log \\
    \midrule
    0.05 &0.049& & 0.147& &2.064& & 0.077& \\
    0.08  &0.049& &0.143& &2.045& & 0.076&  \\
    0.10   &0.048& & 0.142 & &2.050& &0.076&   \\
    0.12  &0.049& &0.144& &2.062& &0.077& \\
    0.15 &0.049 & &0.145& &2.064& &0.077&  \\    
    \bottomrule
  \end{tabular}
  \vspace{2mm}
  \caption{Evaluation of the hyper-parameter $\lambda_{\text{cls}}$ in Equation (10).}
  \label{tab:lambda}
\end{table}

\section{Qualitative Results}
\label{viz}
\begin{figure*}
    \centering
    \includegraphics[width=\linewidth]{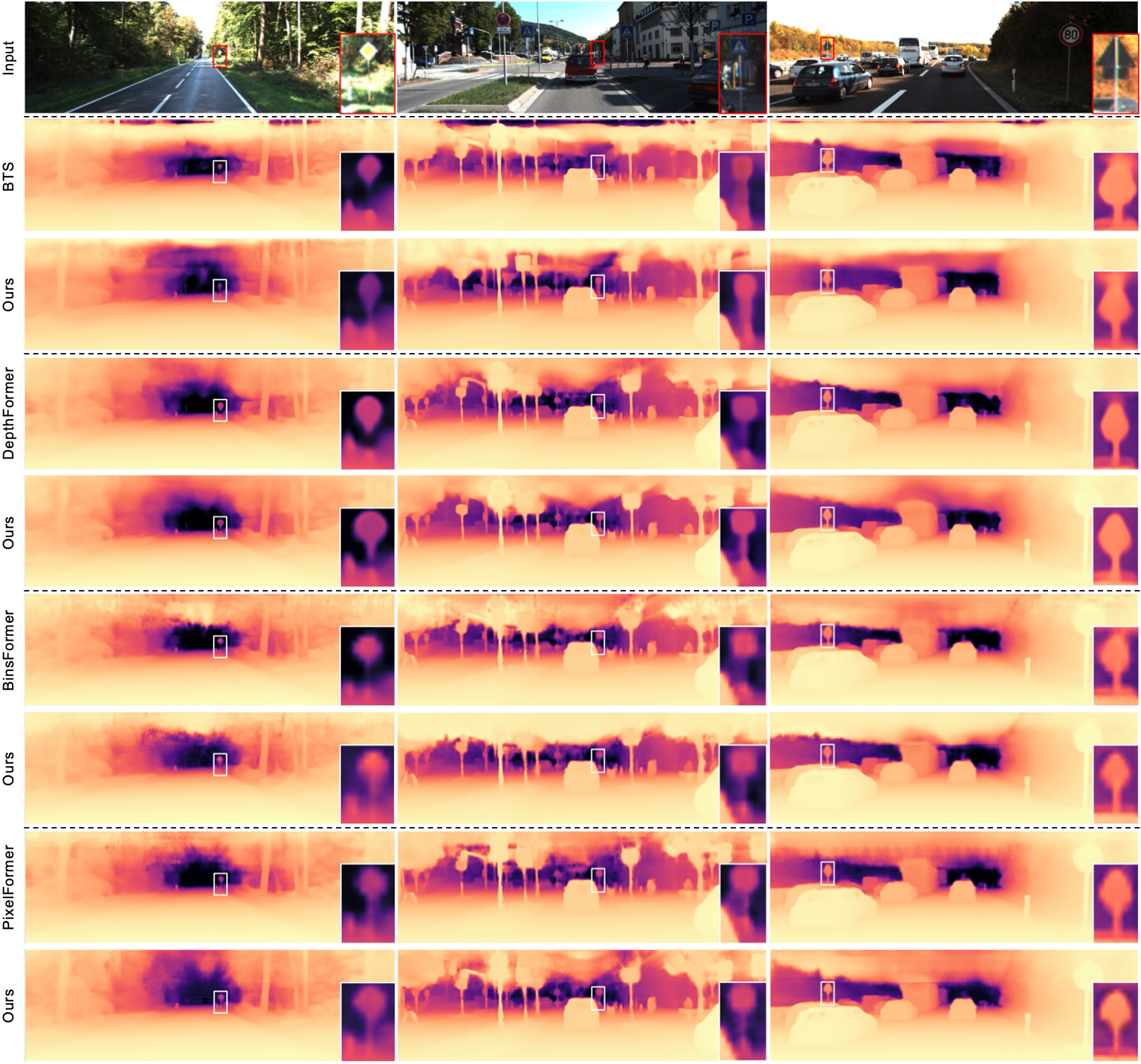}
    \caption{Comparison of the depth prediction results by the state-of-the-art methods and our approach on three scenes of KITTI.}  
    \label{fig:kitti_eigen_qual_sup} 
\end{figure*}

Figure~\ref{fig:kitti_eigen_qual_sup} provides the qualitative results of our approach (GEDepth-Adaptive) and the corresponding state-of-the-art methods on the test set of KITTI. As can be seen in this figure, our approach produces sharper depth prediction, and we observe that the improvements on the distant objects and fine structures are more evident. 

\end{document}